\documentclass[runningheads]{llncs}
\usepackage{graphicx}
\usepackage{amsmath,amssymb} 
\usepackage{color}
\usepackage{hyperref}
\usepackage{booktabs}
\usepackage{threeparttable}

\def\bsb{\boldsymbol{b}}

\def\x{\boldsymbol{\mathrm{x}}}

\def\s{\boldsymbol{\mathrm{s}}}

\def\X{\boldsymbol{\mathrm{X}}}

\def\a{\mathbf{a}}

\def\y{\boldsymbol{\mathrm{y}}}

\def\W{\boldsymbol{\mathrm{W}}}

\def\Re{\mathbb{R}}

\def\bp{\boldsymbol{\mathrm{p}}}

\begin{document}
\title{Fine-grained Video Categorization with Redundancy Reduction Attention} 

\titlerunning{Fine-grained Video Categorization with RRA}
%
\author{Chen Zhu\inst{1}\orcidID{0000-0002-3103-8752} \and
Xiao Tan\inst{2}\orcidID{0000-0001-9162-8570} \and Feng Zhou\inst{3}\orcidID{0000-0002-1132-5877} \and Xiao Liu\inst{2}\orcidID{0000-0002-5689-9786} \and Kaiyu Yue\inst{2}\orcidID{0000-0002-1820-3223} \and Errui Ding\inst{2}\orcidID{0000-0002-1867-5378} \and
Yi Ma\inst{4}}
%
\authorrunning{C. Zhu \and X. Tan \and F. Zhou \and X. Liu \and K. Yue \and E. Ding \and Y. Ma}
%

\institute{University of Maryland, College Park \\
\email{chenzhu@cs.umd.edu} \and
Department of Computer Vision Technology (VIS), Baidu Inc., Beijing \and
Baidu Research, Sunnyvale \\
\email{\{tanxiao01,zhoufeng09,liuxiao12,yuekaiyu,dingerrui\}@baidu.com} \and
University of California, Berkeley\\
\email{yima@eecs.berkeley.edu}}

\maketitle              
\begin{abstract}
For fine-grained categorization tasks, videos could serve as a better source than static images as videos have a higher chance of containing discriminative patterns. 
Nevertheless, a video sequence could also contain a lot of redundant and irrelevant frames. 
How to locate critical information of interest is a challenging task. 
In this paper, we propose a new network structure, known as Redundancy Reduction Attention (RRA), which learns to focus on multiple discriminative patterns by suppressing redundant feature channels. 
Specifically, it firstly summarizes the video by weight-summing all feature vectors in the feature maps of selected frames with a spatio-temporal soft attention, and then predicts which channels to suppress or to enhance according to this summary with a learned non-linear transform. 
Suppression is achieved by modulating the feature maps and threshing out weak activations. The updated feature maps are then used in the next iteration. 
Finally, the video is classified based on multiple summaries. The proposed method achieves outstanding performances in multiple video classification datasets. 
Furthermore, we have collected two large-scale video datasets, YouTube-Birds and YouTube-Cars, for future researches on fine-grained video categorization. 
The datasets are available at \href{http://www.cs.umd.edu/~chenzhu/fgvc}{\url{http://www.cs.umd.edu/~chenzhu/fgvc}}.
\end{abstract}
\section{Introduction}
\begin{figure}[t]
\centering
\includegraphics[width=0.5\linewidth]{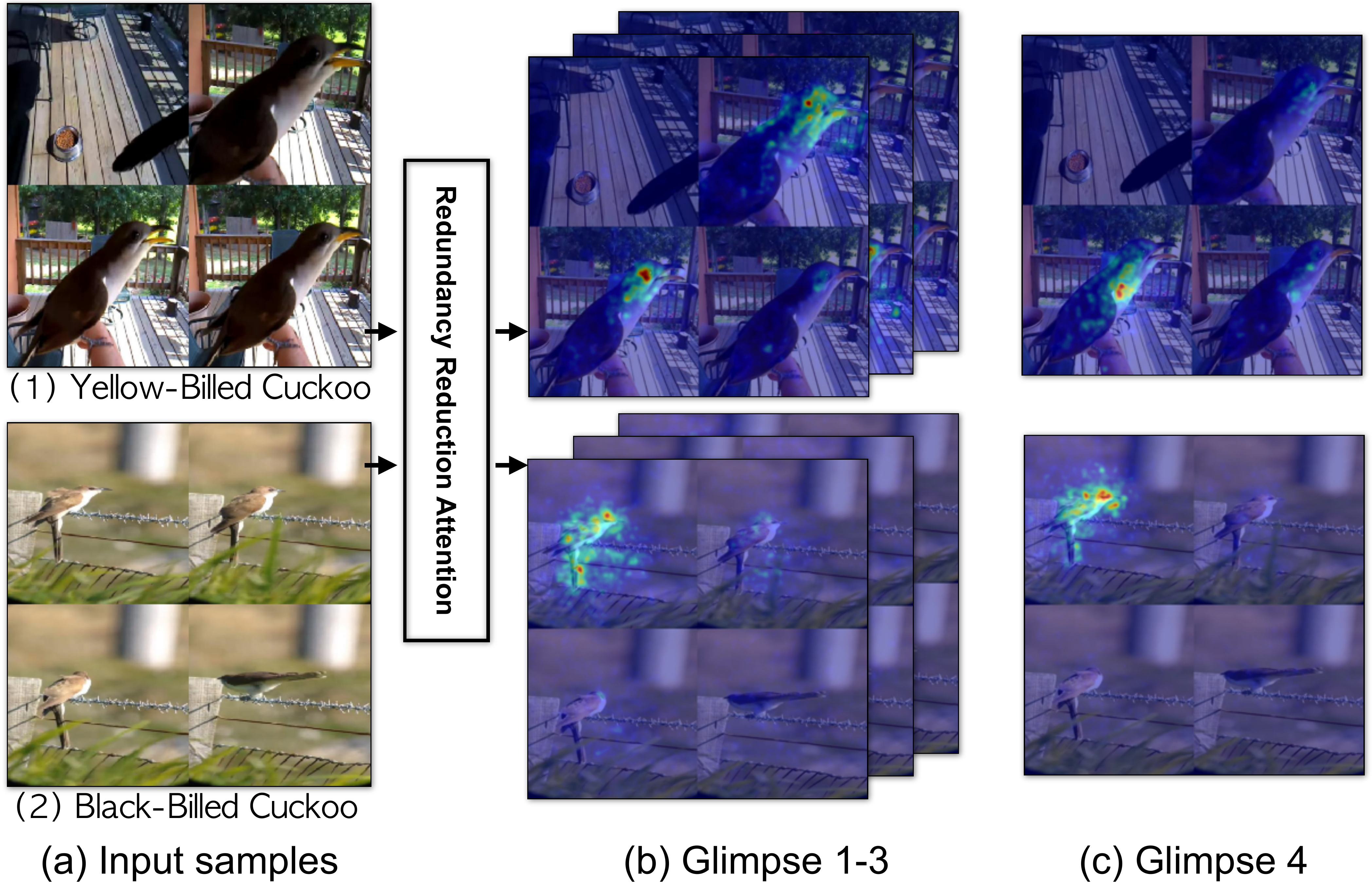}
\caption{Visualization of two real cases on our YouTube-Birds validation set with our RRA model. The heat maps are computed with Equation~\ref{eq:l1} which represents the model's attention on the pixels. This instance has 4 sampled frames and 4 glimpses. Glimpses 2 and 3 are hidden to save space. 
The target in the input frames for the network may be missing or deformed after preprocessing, as in (1) and (2). Our model counters such problems by: 1) Focusing on most discriminative locations among all input frames with soft attention, which helps (1) to ignore the ``empty'' frame. 2) Iteratively depressing uninformative channels, which helps (2) to correct the mis-recognition to House Wren in glimpses 1-3 due to deformation, and recognize correctly with discriminative patterns (head) in glimpse 4. }
\label{fig:teaser}
\end{figure}
Fine-grained visual recognition, such as recognizing bird species~\cite{WahCUB_200_2011,Zheng_2017_ICCV} and car models~\cite{krause20133d,Gebru_2017_ICCV}, has long been of interest to computer vision community. In such tasks, categories may differ only in subtle details, e.g., Yellow-billed Cuckoo and Black-billed Cuckoo, collected in the popular benchmark CUB-200-2011~\cite{WahCUB_200_2011}, look almost the same except for the color of their bills and the patterns under their tails. Hence, lots of works emphasize the importance of discriminative patterns, adopting part annotations~\cite{zhang2016spda,zhang2014part} and attention mechanisms~\cite{Fu_2017_CVPR,Zheng_2017_ICCV}. Progress has been evident on existing datasets, but photos reflecting Cuckoos' bill color or their tail are not always easy to take, as birds seldom keep still and move fast. The discriminative patterns may also become insignificant during the preprocessing process, as shown in Fig.~\ref{fig:teaser}. Recognizing such non-discriminative images is an ill-posed problem.
Instead, videos usually come with abundant visual details, motions and audios of their subjects, which have a much higher chance of containing discriminative patterns and are more suitable than single images for fine-grained recognition in daily scenarios. Nevertheless, videos have higher temporal and spatial redundancy than images. The discriminative patterns of interest are usually present only in a few frames and occupy only a small fraction of the frames. Other redundant frames or backgrounds may dilute the discriminative patterns and cause the model to overfit irrelevant information. 

In this work, we propose a novel neural network structure, called Redundancy Reduction Attention (RRA), to address the aforementioned redundancy problem. It is inspired by the observation that different feature channels respond to different patterns, and learning to reduce the activations of non-discriminative channels leads to substantial performance improvement~\cite{hu2017squeeze,Zheng_2017_ICCV}. In the same spirit, we allow our model to learn to reduce the redundancy and to focus on discriminative patterns by weakening or even blocking non-discriminative channels. Specifically, the model summarizes and updates the feature maps of all input frames iteratively. In each iteration, a soft attention mask is applied over each feature vector of all input feature maps to weight-sum the feature maps into a summary feature vector, and then a learned non-linear transform predicts the increment or decrement of each channel according to the summary feature vector. The increment or decrement is replicated spatially and temporally to each feature vector in the feature maps, and a BN-ReLU block will re-weight and threshold the modified feature maps. With such structures, our model learns to focus on discriminative local features through soft attention while ignoring redundant channels to make each glimpse~\footnote{Refers to $\hat{\x}$ in Eq.~\ref{eq:att}, similar to~\cite{larochelle2010learning}} informative.

Because existing fine-grained video datasets are small~\cite{saito2016ibc127} or weakly-labeled~\cite{karpathy2014large}, we have collected two new large video datasets to remedy for the lack of better fine-grained video datasets. The two datasets are for fine-grained bird species and car model categorization, and are named YouTube Birds and YouTube Cars, respectively. As their names indicate, the videos are obtained from YouTube. They share the same taxonomy as CUB-200-2011 dataset~\cite{WahCUB_200_2011} and Stanford Cars dataset~\cite{krause20133d}, and are annotated via crowd sourcing. YouTube-Cars has 15220 videos of 196 categories, and YouTube-Birds has 18350 videos of 200 categories. To the best of our knowledge, our two datasets are the largest fine-grained video datasets with clean labels.

To sum up, the main contributions of this work are: 1) Proposing a novel redundancy reduction attention module to deal with the redundancy problems in videos explicitly. 2) Collecting two published fine-grained video categorization datasets.  3) Achieving state-of-the-art results on ActivityNet~\cite{caba2015activitynet}, Kinetics~\cite{kay2017kinetics}, as well as our newly collected datasets.




\section{Related Works}
\subsection{Fine-grained Visual Categorization}
State-of-the-art fine-grained categorization approaches mostly employ deep convolutional networks pretrained on ImageNet to extract image features. Some works seek for increasing the capacity of the features e.g., the popular bilinear features~\cite{lin2015bilinear} and recently proposed polynomial kernels~\cite{Cai_2017_ICCV} resort to higher-order statistics of convolutional activations to enhance the representativeness of the network. Despite its success, such statistics treat the whole image equally. There are other methods trying to explicitly capture the discriminative parts. Some of them leverage the manual annotations of key regions~\cite{WahCUB_200_2011,zhang2016spda,zhang2014part} to learn part detectors to help fine-grained classifiers, which requires heavy human involvements. In order to get rid of the labor intensive procedure, attention mechanism is deployed to highlights relevant parts without annotations, which boosts subsequent modules. A seminal work called STN~\cite{jaderberg2015spatial} utilizes localization networks to predict the region of interest along with its deformation parameters such that the region can be more flexible than the rigid bounding box. \cite{Fu_2017_CVPR} improves STN by adopting multiple glimpses to gradually zoom into the most discriminative region, but refining the same region does not fully exploit the rich information in videos. 
MA-CNN~\cite{Zheng_2017_ICCV} learns to cluster spatially-correlated feature channels, localize and classify with discriminative parts from the clustered channels.

\subsection{Video Classification}
It has been found that the accuracy of video classification with only convolutional features of a single frame is already competitive~\cite{karpathy2014large,qiu2017learning}. 
A natural extension to 2D ConvNets is 3D ConvNets~\cite{ji20133d} that convolves both spatially and temporally. P3D ResNet~\cite{qiu2017learning} decomposes a 3D convolution filter into the tensor product of a temporal and a spatial convolution filter initialized with pre-trained 2D ConvNets, which claims to be superior to previous 3D ConvNets. I3D~\cite{carreira2017quo} inflates pretrained 2D ConvNets into 3D ConvNets, achieving state-of-the-art accuracies on major video classification datasets. RNNs is an alternative to capture dependencies in the temporal dimension~\cite{li2017videolstm,sun2017lattice}.

Many of the best-performing models so far adopt a two-stream ensembling~\cite{simonyan2014two}, which trains two networks on the RGB images and optical flow fields separately, and fuse the predictions of them for classification. TSN~\cite{wang2016temporal} improves~\cite{simonyan2014two} by fusing the scores of several equally divided temporal segments.

Another direction is to consider the importance of regions or frames. Attentional Pooling~\cite{girdhar2017attentional} interprets the soft-attention-based classifier as a low-rank second order pooling. 
Attention Clusters~\cite{long2018attention} argues that integrating a cluster of independent local glimpses is more essential than considering long-term temporal patterns.
\cite{zhu2016key} proposes a key volume mining approach which learns to identify key volumes and classify simultaneously. 
AdaScan~\cite{Kar_2017_CVPR} predicts the video frames' discrimination importance while passing through each frame's features sequentially, and computes the importance-weighted sum of the features. \cite{sharma2015action} utilizes a 3-layer LSTM to predict an attention map on one frame at each step. The afore-mentioned two methods only use previous frames to predict the importance or attention and ignore the incoming frames. In addition, all methods mentioned above lack of a mechanism which can wisely distinguish the informative locations and frames in videos jointly. 
To be noted, Attend and Interact~\cite{ma2017attend} considers the interaction of objects, while we focus on extracting multiple complementary attentions by suppressing redundant features.




\section{Methods}
\begin{figure*}[htbp]
\centering
\includegraphics[width=\linewidth]{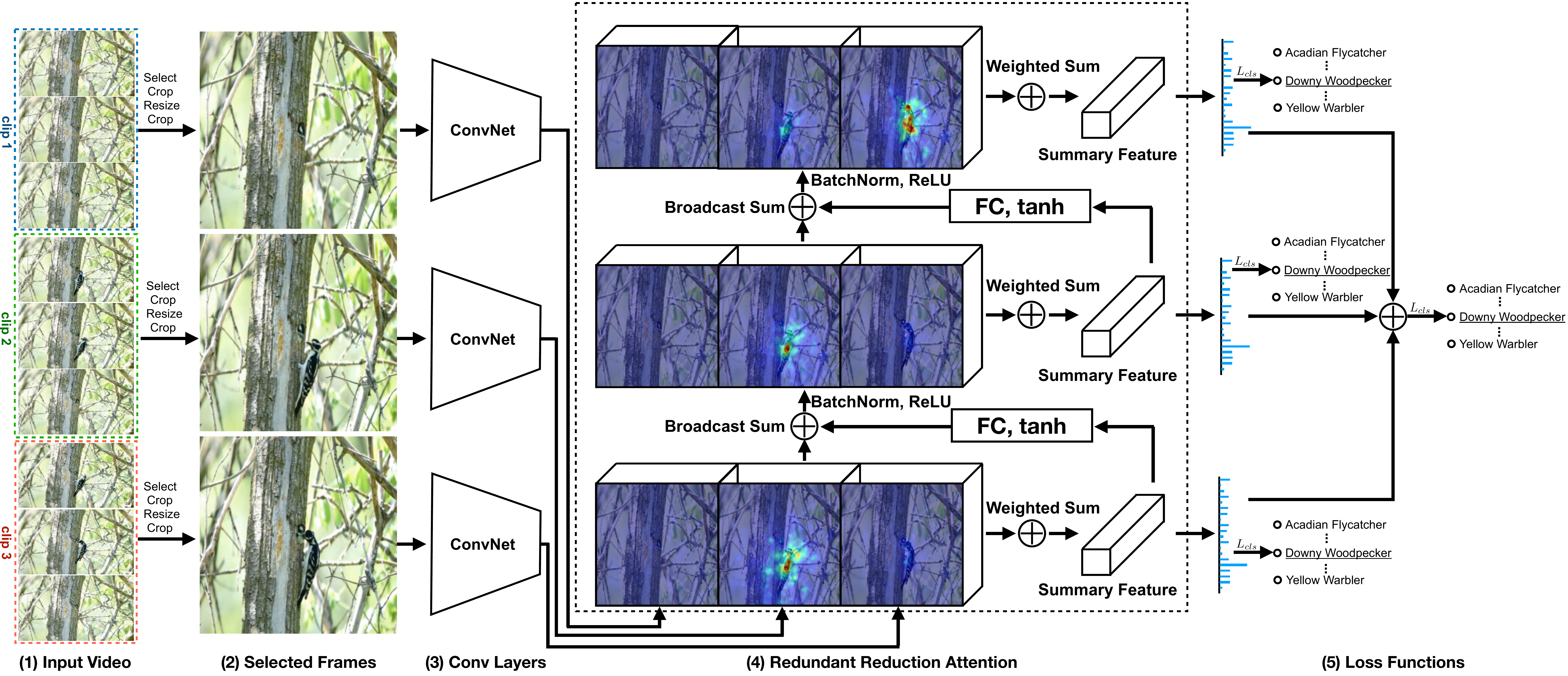}
\caption{The general structure of the proposed model. Input sequences are divided into clips of the same length. One frame or flow stack is sampled from each clip. The CNNs extract feature maps from the sampled frames, then the RRA modules iteratively updates the feature maps. Each summary feature vector gives one classification score via the classifiers, and the scores are averaged as the final prediction. }
\label{fig:general}
\end{figure*}

Figure \ref{fig:general} shows the overall structure of the proposed network. The same structure can be used to handle both RGB and optical flow inputs, except for changing the first convolution layer to adapt to stacked optical flows. Generally, our model learns to focus on the most discriminative visual features for classification through soft attention and channel suppression. For the inputs, we take a frame from each uniformly sliced temporal clip to represent the video. For training, each clip is represented by a random sample of its frames to increase variety of training data. For testing, frames are taken at the same index of each clip.
Before going into details, we list some notations to be used throughout the paper. Denote the width and the height of feature maps as $w$ and $h$. $\x_i\in \Re^{c\times hw}$ is the convolutional feature map of the $i$-th frame, $\X=[\x_1,...,\x_n]\in \Re^{c\times nhw}$ is the matrix composed of feature maps of all the $n$ frames. $\bar{\X}$ is the redundancy-reduced $\X$ to be described in Section~\ref{sec:rra}. We use $A \oplus B$ to denote the operation of replication followed by an element-wise sum, where the replication transforms $A$ and $B$ to have the same dimensions. The superscript $^{k}$ represents $k$-th iteration.


\subsection{Redundancy Reduction Attention}\label{sec:rra}
\begin{figure}[htbp]
\centering
\includegraphics[width=0.4\linewidth]{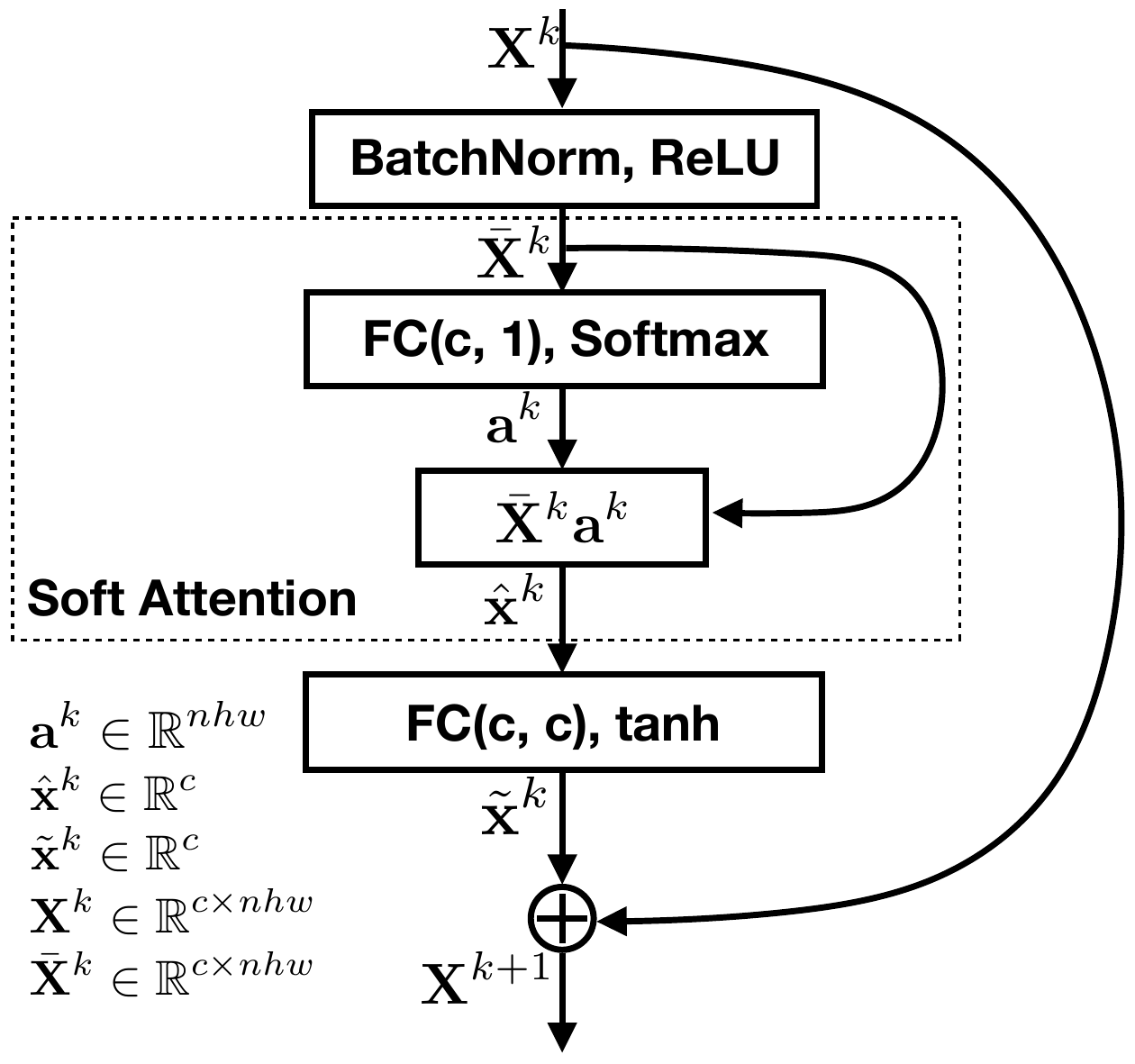}
\caption{Structure of one RRA module. RRA network is constructed by concatenating such modules. The final addition is a broadcasting operator.}
\label{fig:resblock}
\end{figure}

Due to duplication of contents, the spatio-temporal feature representation $\X$ is highly redundant. In this section, we introduce a new network structure shown in Fig.~\ref{fig:resblock} which is able to attend to the most discriminative spatio-temporal features and suppress the redundant channels of feature maps.

The soft attention mechanism~\cite{fukui2016multimodal,xu2015show} is able to select the most discriminative regional features. We extend it to the spatio-temporal domain to infer the most discriminative features of the video for categorization and reduce redundancy. 
As shown in our ablation experiments, unlike the spatial-only attention, it prevents the most discriminative features from being averaged out by background features. The attention weights $\a\in \Re^{nhw}$ are modeled as $\a = \mathrm{softmax}(\bar{\X}^T\W_a)$, 
where $\W_a\in \Re^{c}$ is learnable, and $\bar{\X}$ is defined in Eq.~\ref{eq:xbar}. The feature vectors of $\X$ are then weight-summed by $\a$ to get the summary vector:
\begin{equation}\label{eq:att}
	\hat{\x} = \bar{\X}\a.
\end{equation}

\begin{figure}[t]
\centering
\includegraphics[width=0.8\linewidth]{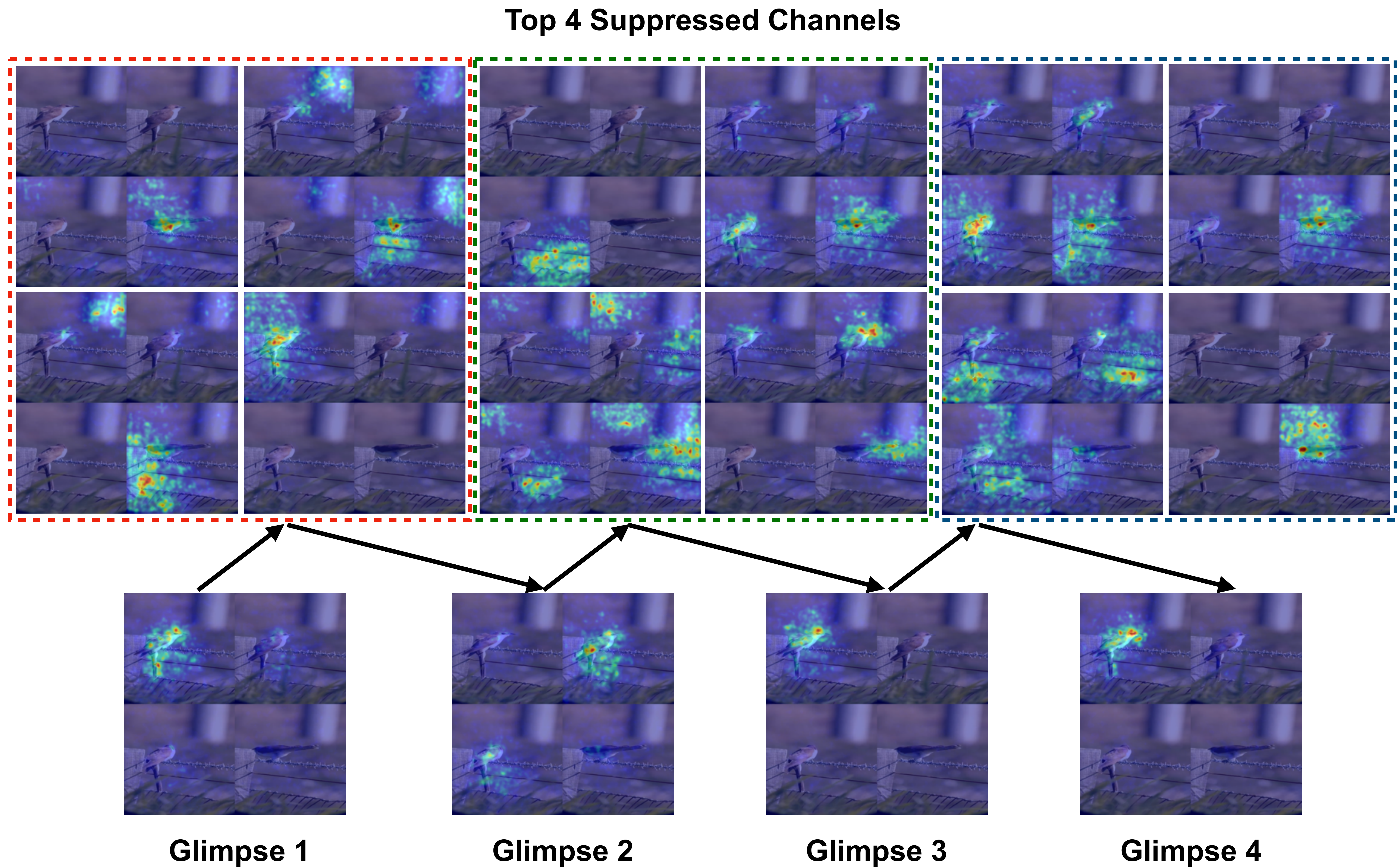}
\caption{One instance of redundancy suppression. Input frames are the same as Figure \ref{fig:teaser}. The top four suppressed channels are selected as the smallest four entrys' indices in $\tilde{\x}^k$, which are channels given the most decrements. We then compute $I_{vis}$ in Section \ref{sec:visualize} by setting $a_i$ as all decreased entries from $\X^k$ to $\bar{\X}^{k+1}$ in these channels, and setting $w_i$ as their respective decrements. The suppressions does not overlap with the next target, and are on meaningful patterns. Red colors indicate higher suppression. }
\label{fig:suppress}
\end{figure}

Since videos contain rich context for classification, it is natural to think of extracting multiple discriminative features with multiple attentions. 
However, we do not want the summaries to duplicate. 
We herein introduce a simple but effective approach which iteratively suppresses redundant feature channels while extracting complementary discriminative features, named Redundancy Reduction Attention (RRA). 
By reduction we refer to decreasing the magnitude of the activations. 
In $k$-th step, the channel-wise reduction $\tilde{\x}^k$ is inferred from the non-linear transform of the summary $\hat{\x}^k$. 
In the case of Fig.~\ref{fig:resblock}, the non-linear transform is selected as a fully connected layer followed by a tanh activation. Reduction is achieved by adding $\tilde{\x}^k$ to the ReLU activation feature map $\X^k$, which is further augmented by the BatchNorm-ReLU~\cite{ioffe2015batch} block to threshold out activations below the average to get the redundancy-reduced feature map $\bar{\X}^{k+1}$: 
\begin{equation}\label{eq:xbar}
\bar{\X}^{k+1}=\mathrm{ReLU}(\mathrm{BatchNorm}(\X^{k}\oplus\tilde{\x}^k))
\end{equation}
Since the range of $\tilde{\x}^k$ is $(-1,1)$, $\tilde{\x}^k$ can not only suppress redundant channels but also enhance the informative channels to produce a more preferable feature map $\X^{k+1}$. As demonstrated in the experiments, using tanh as the activation for~$\tilde{\x}^{k}$ is better than the $-\mathrm{ReLU}(x)$ alternative. 
A visualization of the suppression process is shown in Figure~\ref{fig:suppress}. 


\subsection{Loss Functions}\label{sec:loss}
We utilize a Softmax classifier to predict the video's label distribution $\hat{\y}$ from the summary feature $\hat{\x}$ as $\hat{\y}=\mathrm{softmax}(\W_c\hat{\x}+\bsb_c)$. A cross entropy loss is applied to minimize the KL divergence between the ground truth distribution $\y$ and $\hat{\y}$:
\begin{equation}\label{eq:ce_loss}
L(\hat{\y}, \y) = -\sum_{i}\y_i\log \hat{\y}_i
\end{equation}
For models with more than one RRA module (iterations), fusing the summary vectors for classification is a natural choice. We have explored three approaches to achieve the fusion.

\textbf{Concatenation Loss $L_c$}: Equivalent to the multi-glimpse models such as~\cite{fukui2016multimodal} which concatenates the glimpse features into a higher dimensional feature vector, we compute each glimpse score $\s^k=\W_c^k\hat{\x}^k+\bsb_c^k$ first, and minimize the cross entropy loss $L_c=L(\hat{\y}_{cat}, \y)$ of their sum
\begin{equation}\label{eq:concat_loss}
\hat{\y}_{cat} = \mathrm{softmax}(\sum_{k=1}^{K} \s^k).\\
\end{equation}
This approach is broadly used, but since the scores are not normalized, they do not necessary have the same scale. If one glimpse gives extremely high magnitude, then other glimpses will be drowned, and the softmax loss may also reach saturation where the gradient vanishes, which harms the performance. In our experiments, we also find this loss suboptimal.

\textbf{Individual Loss $L_i$}: To overcome the normalization problem of $L_c$, we directly supervise on each of the individual glimpses. That is, we can apply cross entropy loss on each glimpse's categorical distribution $\hat{\y}^k$ and minimize their sum,
\begin{equation}\label{eq:individual_loss}
L_i=\sum_{k=1}^{K} L(\hat{\y}^k, \y).
\end{equation}
This loss and its combinations perform the best in our experiments.

\textbf{Ensemble Loss $L_e$}: Since we have actually trained several classifiers with $L_i$, we could ensemble results from different glimpses as
\begin{equation}\label{eq:ensemble_score}
\bar{\y} = \frac{1}{K}\sum_{k=1}^{K}{\hat{\y}^k},
\end{equation}
and compute $L_e=L(\bar{\y},\y)$.
This is in fact optimizing the ensemble score directly. In our experiments, this loss does not perform well alone, but improves the performance when combined with other losses. 

The losses can be summed to achieve different objectives. Although not explored in this paper, weights can also be applied on each loss, and even as trainable parameters reflecting the importance of each glimpse when computing $L_e$ and the final scores.

\subsection{Visualizing Attention over the Input}\label{sec:visualize}
To check whether the network has really learned to focus on discriminative parts, we visualize each pixel's influence on the distribution of attention $\a$. Since $||\a||_1=1$, $L_{vis}=\frac{1}{2}||\a||_2^2$ reflects $\a$'s difference from mean pooling. We expect its distribution to highlight the discriminative patterns, which is probably far from mean pooling. Further, its derivative w.r.t. a input pixel $\bp\in \Re^3$ is $\frac{\partial L_{vis}}{\partial \bp} =\sum_{i=1}^{nhw} \frac{\partial L_{vis}}{\partial a_i}\frac{\partial a_i}{\partial \bp}=\sum_{i=1}^{nhw} w_i\frac{\partial a_i}{\partial \bp}$
where $w_i=a_i$. It not only reflects $\bp$'s influence on $a_i$ with $\frac{\partial a_i}{\partial \bp}$, but also reflects how much attention is paid to this influence by the weight $w_i$. With this equation, we can also set $w_i$ to other values to weigh the influences. Finally, we quantize the attention-weighed influence by the $\ell^1$ norm of this derivative
\begin{equation}\label{eq:l1}
I_{vis}=\left\lVert\frac{\partial L_{vis}}{\partial \bp}\right\lVert_1,
\end{equation}
and use a color map on $I_{vis}$ to enhance the visual difference. A Gaussian filter is applied to make high values more distinguishable.

\section{Novel Fine-grained Video Datasets}

In order to provide a good benchmark for fine-grained video categorization, we built two challenging video datasets, YouTube Birds and YouTube Cars, which consist of 200 different bird species and 196 different car models respectively. The taxonomy of the two datasets are the same as CUB-200-2011~\cite{WahCUB_200_2011} and Stanford Cars~\cite{krause20133d} respectively. Fig.~\ref{fig:dataset} shows some sample frames from the two datasets.
Compared with the two reference datasets, subjects in our datasets have more view point and scale changes. YouTube Birds also doubles the size of IBC127~\cite{saito2016ibc127}, a video dataset with 8,014 videos and 127 fine-grained bird categories.
Table~\ref{tab:dataset_specs} lists the specifications of the annotated datasets. $N_c$ is number of categories. $N_{train}$ and $N_{test}$ are number of training and testing videos. $n_v$ and $m_v$ are minimum and maximum number of videos for a category.
\begin{table}[htbp]
\begin{minipage}[b]{0.65\linewidth}
\centering
\includegraphics[width=\linewidth]{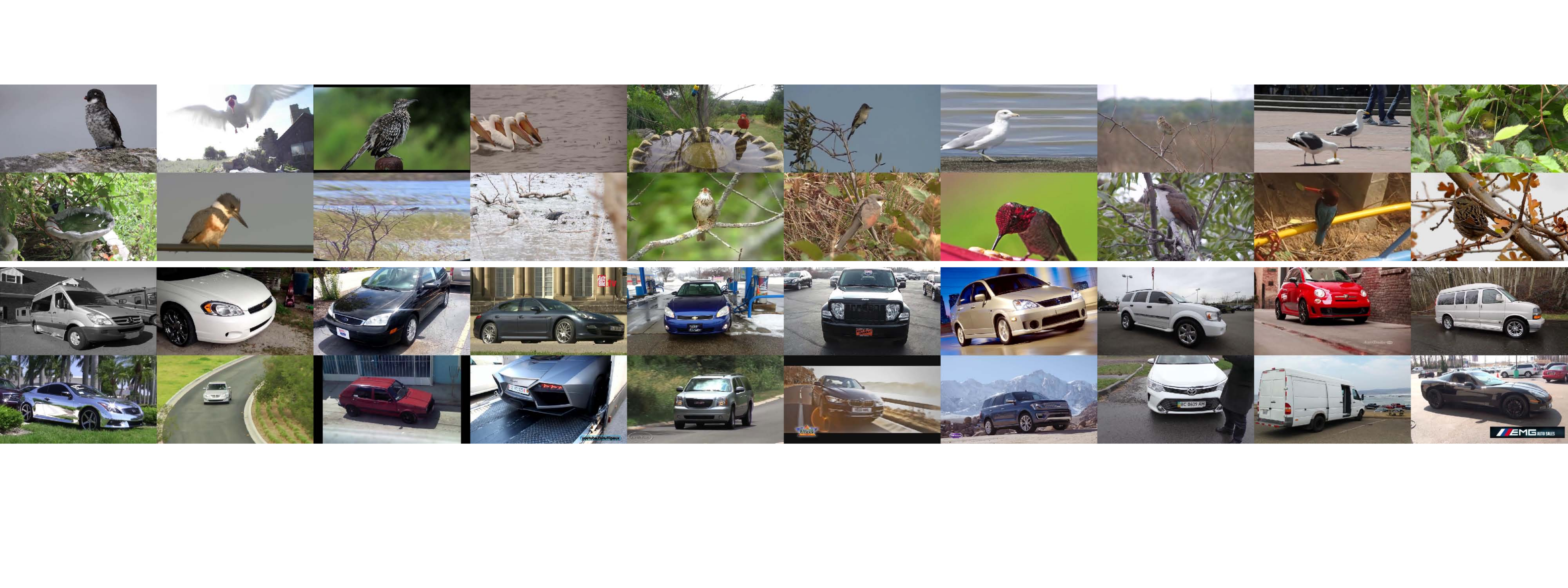}
\caption{Sample frames from YouTube Birds and YouTube Cars datasets. Top 2 rows are from YouTube Birds, bottom 2 rows are from YouTube Cars.}
\label{fig:dataset}
\end{minipage}
\begin{minipage}[b]{0.34\linewidth}
  \centering
  \makebox[\linewidth]{\resizebox{\linewidth}{!}{
      \begin{threeparttable}
        \begin{tabular}{cccccc}
          \toprule
          Set           & $N_c$ & $N_{train}$ & $N_{test}$  &  $n_v$ & $m_v$  \\
          \midrule
          Birds  & 200   & 12666 & 5684  &    6  &  249     \\
          Cars   & 196   & 10259 & 4961  &    6  &  207     \\
          \bottomrule
        \end{tabular}
      \end{threeparttable}
    }}
  \caption{Specifications of YouTube Birds and YouTube Cars. }
  \label{tab:dataset_specs}
\end{minipage}
\end{table}

Videos of both datasets were collected through YouTube video search. We limited the resolution of videos to be no lower than 360p and the duration to be no more than 5 minutes.
We used a crowd sourcing system to annotate the videos. Before annotating, we firstly filter the videos with bird and car detectors to ensure at least one of the sample frames contains a bird or a car. For each video, the workers were asked to annotate whether each of its sample frames (8 to 15 frames per video) belong to the presumed category by comparing with the positive images (10 to 30 per category) of that category. 
As long as there is one sample frame from the video belong to the presumed category, the video will be kept.
According to the annotations, about 29\% and 50\% of the frames of YouTube Birds/YouTube Cars contain a bird/car.
However, since one video may contain multiple subjects from different categories, there may be more than one category in the same video. To make evaluation easier, we removed all videos appearing in more than one category. Videos of each category were split into training and test sets in a fixed ratio. More details are in the project page.

\section{Experimental Results}



\begin{figure*}
\centering
\includegraphics[width=\linewidth]{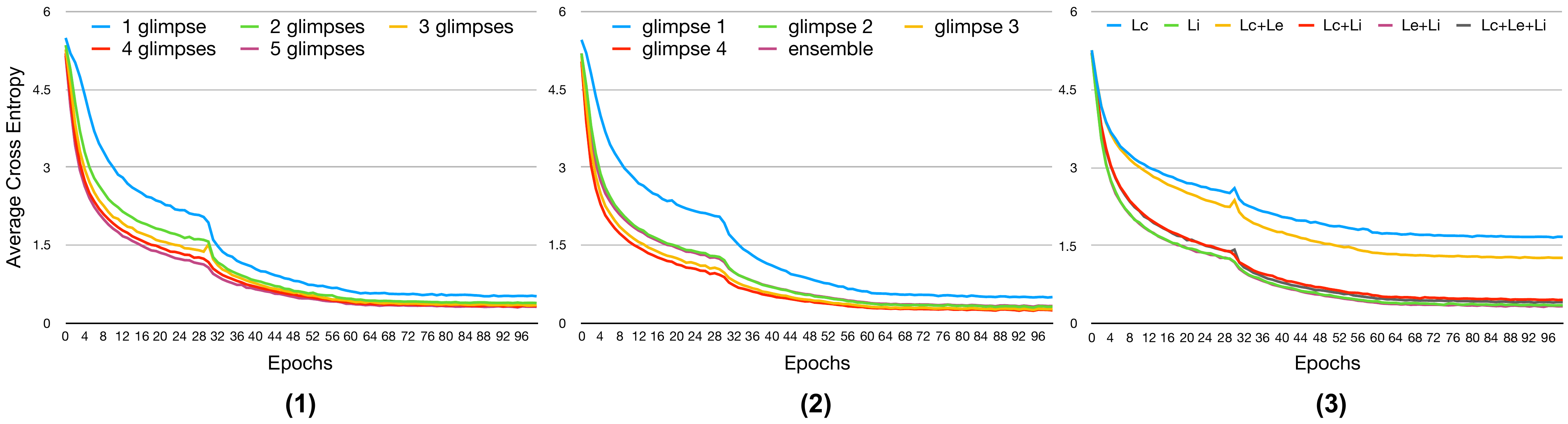}
\caption{Average loss curves throughout epochs on ActivityNet v1.3 training set. (1): loss curves w.r.t. different number of glimpses. As the number of glimpse increases, it converges quicker, and indicates better generalization on validation set. (2): loss curves of each glimpse and the ensemble score in the 4-glimpse model with only $L_i$. (3): loss curves of different loss functions. The $L_e$ curve is ignored - the curve is ascending.}
\label{fig:loss_curve}
\end{figure*}
We evaluated the proposed method for general video categorization and fine-grained video categorization. For general tasks, we selected activity recognition and performed experiments on RGB frames of ActivityNet v1.3~\cite{caba2015activitynet} and both RGB and flow of Kinetics \cite{kay2017kinetics}. For fine-grained tasks, we performed experiments on our novel datasets YouTube Birds and YouTube Cars.\par
We first introduce the two public datasets and our experimental settings, and then analyze our model with controlled experiments. Finally we compare our method with state-of-the-art methods.

\begin{table}[htbp]
  \centering
  \makebox[\linewidth]{\resizebox{0.5\linewidth}{!}{
      \begin{threeparttable}
        \begin{tabular}{cccccc}
          \toprule
          Loss      &   $\mathrm{mAP}_c$ &  $\mathrm{mAP}_e$  & Loss        & $\mathrm{mAP}_c$ &  $\mathrm{mAP}_e$ \\
          \midrule
          $L_c$     &   80.27            & 77.84              &  $L_i$      &    82.60         &   82.97 \\
          $L_e$     &   25.75            & 36.24              &  $L_c+L_i$  &    82.41        &   82.80\\
          $L_c+L_e$ &   81.48            & 80.45              &  $L_e+L_i$  &    82.90        &    \textbf{83.42}  \\
          $L_c+L_i+L_e$ & 82.28          & 82.59              &     -       &      -          & -\\
          \bottomrule
        \end{tabular}
      \end{threeparttable}
    }}
  \caption{Ablation analysis of loss functions on ActivityNet v1.3 validation set. $\mathrm{mAP}_e$ stands for the mAP of ensemble score, $\mathrm{mAP}_c$ stands for the mAP of concatenation score.}
  \label{tab:untrimmed_loss}
\end{table}

\subsection{Settings}
\textbf{ActivityNet v1.3}~\cite{caba2015activitynet}: It has 200 activity classes, with 10,024/4,926/5,044 training/validation/testing videos. Each video in the dataset may have multiple activity instances. There are 15,410/7,654 annotated activity instances in the training/validation sets respectively. The videos were downsampled to 4fps. We trained on the 15,410 annotated activity instances in the training set, and kept the top 3 scores for each of the 4,926 validation videos. We report the performances given by the official evaluation script.

\textbf{Kinetics}~\cite{kay2017kinetics}: This dataset contains 306,245 video clips with 400 human action classes. Each clip is around 10 seconds, and is taken from different YouTube videos. Each class has 250-1000 clips, 50 validation clips and 100 testing clips. The optical flows were extracted using TV-L1 algorithm implemented in OpenCV. We did not downsample the frames on this dataset. The results were tested with official scripts on the validation set.


\textbf{YouTube Birds and YouTube Cars}: We only experiment on the RGB frames of the 2 datasets. Videos in YouTube Birds and YouTube Cars were downsampled to 2fps and 4fps respectively. We split the datasets as in Table~\ref{tab:dataset_specs}.

\textbf{Training:} We trained the model in an end-to-end manner with PyTorch. The inputs to our model are the label and 4 randomly sampled RGB frames or flow stacks (with 5 flow fields) from 4 equally divided temporal segments. We adopted the same multi-scale cropping and random flipping to each frame as TSN for data augmentation. We used ImageNet pretrained ResNet-152~\cite{he2016deep} provided by PyTorch and ImageNet pretrained Inception-V3~\cite{szegedy2016rethinking} provided by Wang et al.~\cite{wang2016temporal} for fair comparisons. We used Adam~\cite{kingma2014adam} optimizer, with an initial learning rate 0.0002 and a learning rate decay factor 0.1 for both RGB and flow networks. Batch size is set to 256 on all datasets. For ActivityNet, YouTube Birds and YouTube Cars, we decayed the learning rate every 30 epochs and the total number of epochs was set to 120, while on Kinetics, we decayed learning rate every 13000 and 39000 iterations for RGB and flow networks respectively. The pretrained convolutional layers were frozen until 30 epochs later on ActivityNet, YouTube Birds and YouTube Cars, and 5 epochs later on Kinetics. Dropout is added before each classification FC layer and set to 0.7/0.5 for RGB/flow respectively.

\textbf{Testing:}
We followed the standard TSN testing protocol, where each video was divided into 25 temporal segments.
One sample frame was taken from the middle of each temporal segment, and the sample was duplicated into 5 crops (top-left, top-right, bottom-left, bottom-right, center) in 2 directions (original + horizontal flipping), i.e., inputs were 250 images for each video. 

\subsection{Ablation Studies}
First, we evaluated the performance of RRA model on ActivityNet v1.3 with different loss functions as proposed in Section~\ref{sec:loss}.
We enumerated all possible combinations of the 3 losses.
For combinations with more than one loss, all losses are equally weighted.
All variants used ResNet-152 as the base network, and were configured to have 4 glimpses.
Table \ref{tab:untrimmed_loss} lists the mAP of the concatenation score (Equation \ref{eq:concat_loss}), and the ensemble score (Equation \ref{eq:ensemble_score}).
We can see that when combined with another loss, $L_e$ generally improves the performance. $L_c$, on the contrary, undermines the accuracy when combined with $L_i$ or $L_i+L_e$. However, training with $L_e$ alone does not converge. It is probably because without individual supervision for each glimpse, training all glimpses jointly is difficult to achieve. In addition, since $L_c$ directly supervises on the concatenate score, $L_c$ and $L_c+L_e$ have higher $\mathrm{mAP}_c$ than $\mathrm{mAP}_e$. From the mAP values, we can see that for our model, $L_i$ is the best single loss, and $L_e+L_i$ is the best combination.

Fig.~\ref{fig:loss_curve} (3) shows the average loss of each epoch on the ActivityNet training set with different kinds of losses. We can see that adding $L_e$ does not change the curves of $L_i$ and $L_c+L_i$ so much, though it does improve the performance when added to them. To be noted, $L_i$ achieved top-1 accuracy of 83.03 with frozen BN, a trick used in TSN. However, in our experiments, frozen BN does not improve the $L_e+L_i$ objective.

We also compared our model with parallel glimpses model. A $k$ parallel glimpses model predicts $k$ glimpses and concatenates the summary feature vectors for classification. More glimpses generally improve the performance, which is quite reasonable. And without surprise, our model is better than parallel glimpse models. The best mAP of 4 parallel glimpse model on ActivityNet v1.3 is 82.39, while the mAP our best RRA model is 83.42.


\begin{table}[htbp]
  \centering
   \makebox[0.45\linewidth]{\resizebox{0.45\linewidth}{!}{
     \begin{tabular}{cccccc}
      \toprule
      \#Glimpses &1&2&3&4&5\\
       mAP & 80.89 & 82.14 & 82.12 & \textbf{83.42} & 82.94 \\
      \bottomrule
    \end{tabular}
   }}
    \makebox[0.45\linewidth]{\resizebox{0.45\linewidth}{!}{
        \begin{tabular}{ccccccc}
      \toprule
      No. &1&2&3&4&5&6\\
      mAP & 80.20 & 81.97 & 82.41 & 83.15 & 82.75 & 82.75 \\
      \bottomrule
    \end{tabular}
   }}
  \caption{Ablation mAPs on the ActivityNet v1.3 validation set, with ResNet-152. \emph{Left}: changing number of glimpses from 1 to 5. \emph{Right}: modifying RRA module into: 1.spatio-temporal average pooling instead of attention; 2.spatial attention and temporal average pooling; 3.no BN; 4.no ReLU; 5.no tanh; 6.-ReLU(x) instead of tanh(x). All the settings are the same as the 83.42 mAP model except for the specified variations.}
  \label{tab:ablation}
\end{table}

Second, we evaluated RRA model with different number of glimpses. In this experiment, the base network is ResNet-152, and the loss is $L_i+L_e$.
Fig.~\ref{fig:loss_curve} (1) shows the average training cross entropy of the ensemble score under different number of glimpses. Generally, with more glimpses, it converges more rapidly, and when glimpse number reaches 4, further increase in glimpse number brings much less acceleration in convergence, and the validation mAP starts to drop, as shown in Table~\ref{tab:ablation} (Left). So in most of our experiments, we have set it to 4. Fig.~\ref{fig:loss_curve} (2) shows the cross entropy of each glimpse's individual score, and the cross entropy of ensemble scores, which helps to explain why adding more glimpses accelerates the convergence of the ensemble score. Glimpses at later iterations converge more rapidly,  which indicates redundancy is removed and they have extracted more discriminative features for classification. With more accurate glimpses, the ensemble score also becomes better, hence converging faster. To check the difference between the glimpses, the top-1 accuracy for each glimpse and their ensembling of the 4-glimpse model is 77.49, 79.09, 78.71, 78.92 and 78.81 respectively. 

Third, we evaluate the role of each component in Fig.~\ref{fig:resblock} by removing or changing one of them and validate the mAP on ActivityNet v1.3. The results are shown in Table \ref{tab:ablation} (Right). Attention plays the most important role, without which the mAP drops by 3.22. If replace the spatio-temporal attention with spatial attention and temporal average pooling, the mAP is better than average pooling, but still worse than spatio-temporal attention. The tanh activation is more suitable as the activation for the reduction as replacing it with a linear transform (removing it directly) or -ReLU(x) decreases the mAP by 0.67. Batch normalization and ReLU are also important components.

\subsection{Comparison with State-of-the-arts}
  

After validating the configurations of the model, we fix the loss function as $L_i+L_e$, the number of glimpses to 4, then train and test on our two datasets along with the two action recognition datasets.

Table \ref{tab:activitynet} (left) shows results on ActivityNet v1.3, where the results of state-of-the-art methods all come from published papers or tech reports. With only RGB frames, our network already out competes 3D CNN-like methods, including the recently proposed P3D \cite{qiu2017learning} which uses ImageNet pretrained ResNets to help initialization. To be noted, our model on ActivityNet v1.3 only used 4fps RGB frames for both training and validation due to physical limitations.

We further evaluate our model on the challenging Kinetics dataset with both RGB and optical flow inputs. Table \ref{tab:activitynet} (right)  shows the comparison with state-of-the-art results on Kinetics dataset. Results of 3D ResNet, TSN and ours are on the validation set while I3D is on the test set. Results of TSN come from their latest project page. Our fusion result is achieved by adding RGB and flow scores directly. Our method surpasses TSN on both RGB and optical flow by significant margins, but the fusion result is a bit lower, which might due to sampling the same frames for both RGB and flow at validation. 

\begin{figure*}
\centering
\includegraphics[width=\linewidth]{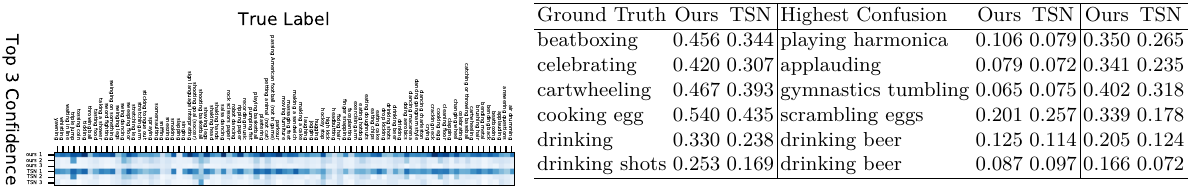}
\caption{\textit{Left}: top-3 confidences for the classes. Darker color indicates higher confidence, and all highest-confidence predictions are correct. \textit{Right}: confidences of the ground truth (first 3 columns) and the most-confusing class (next 3 columns), and the gaps (last 2 columns). Our model's mAP is 73.7 while the TSN's is 72.5. Both models' highest confidence is less than 0.5 in these cases. }
\label{fig:confusion}
\end{figure*}

To demonstrate the reduction of confusion brought by our model, in Fig.~\ref{fig:confusion} we show some of TSN and our model's top-3 average confidences from the confusion matrix on confusing classes of the Kinetics dataset.
Our model has a systematically higher average confidence on the correct classes and a clearer gap between correct and wrong classes.
\begin{table}[htbp]
\begin{minipage}[b]{0.66\linewidth}
  \centering
      \makebox[0.45\linewidth]{\resizebox{0.45\linewidth}{!}{
        \begin{tabular}{cccc}
          \toprule
          Method & top-1 & mAP & top-3    \\
          \midrule
          IDT~\cite{wang2013action} &  64.70 & 68.69 & 77.98 \\
          C3D~\cite{qiu2017learning} & 65.80 & 67.68 & 81.16 \\
          P3D~\cite{qiu2017learning} & 75.12 & 78.86 & 87.71  \\
          Ours & \textbf{78.81}  & \textbf{83.42} & \textbf{91.88}\\
          \bottomrule
        \end{tabular}
      }}
      \makebox[0.45\linewidth]{\resizebox{0.5\linewidth}{!}{
        \begin{tabular}{cccc}
          \toprule
          Method & RGB & Flow & Fusion  \\
          \midrule
          3D ResNet~\cite{hara2017learning} & 58.0 & - & - \\
          I3D~\cite{carreira2017quo}*  & 71.1 & 63.4 & 74.2 \\
          TSN~\cite{wang2016temporal}   & 72.5 & 62.8 & \textbf{76.6} \\
          Ours  & \textbf{73.7} & \textbf{63.9} & 76.1  \\
          \bottomrule
        \end{tabular}
      }}
  \caption{\emph{Left}: Results on the ActivityNet v1.3 validation dataset, with ResNet-152. \emph{Right}: Top-1 accuracies on the Kinetics dataset, with ResNet-152. }
  \label{tab:activitynet}
\end{minipage}
\begin{minipage}[b]{0.33\linewidth}
\centering
  \makebox[1\linewidth]{\resizebox{0.9\linewidth}{!}{
  \begin{threeparttable}
    \begin{tabular}{ccc}
      \toprule
      Method & Birds & Cars \\
      \midrule
      BN-Inception & 60.13 & 61.96 \\
      I3D(Res50)  & 40.68 & 40.92  \\
      TSN~\cite{wang2016temporal}  &  72.361     & 74.340 \\
      Ours &  \textbf{73.205}       & \textbf{77.625} \\
      \bottomrule
    \end{tabular}
  \end{threeparttable}
  }}
  \caption{Comparing with methods on YouTube Birds and YouTube Cars. }
  \label{tab:birds}
\end{minipage}
\end{table}

Finally, Table \ref{tab:birds} shows results on YouTube Birds and YouTube Cars. The BN-Inception model randomly takes one frame from each video during training and takes the middle frame for testing. Similarly, I3D(Res50)~\cite{carreira2017quo} is initialized by inflating an ImageNet-pretrained ResNet-50. It takes 32 consecutive frames at a random time or in the middle of the video for training and testing respectively. For TSN, we use its official implementation in PyTorch and the ImageNet pretrained Inception-V3 model provided by its authors for fair comparison. Our model also used the same Inception-V3 model for initialization. Our method surpasses TSN on these two datasets, since categories in fine-grained tasks often share many features in common and hence require a higher level of redundant reduction and to focus more on the informative locations and frames. A even larger margin is especially evident on YouTube Cars for the similar reason. 

\subsection{Qualitative Results}
\begin{figure*}[htbp]
\centering
\includegraphics[width=\linewidth]{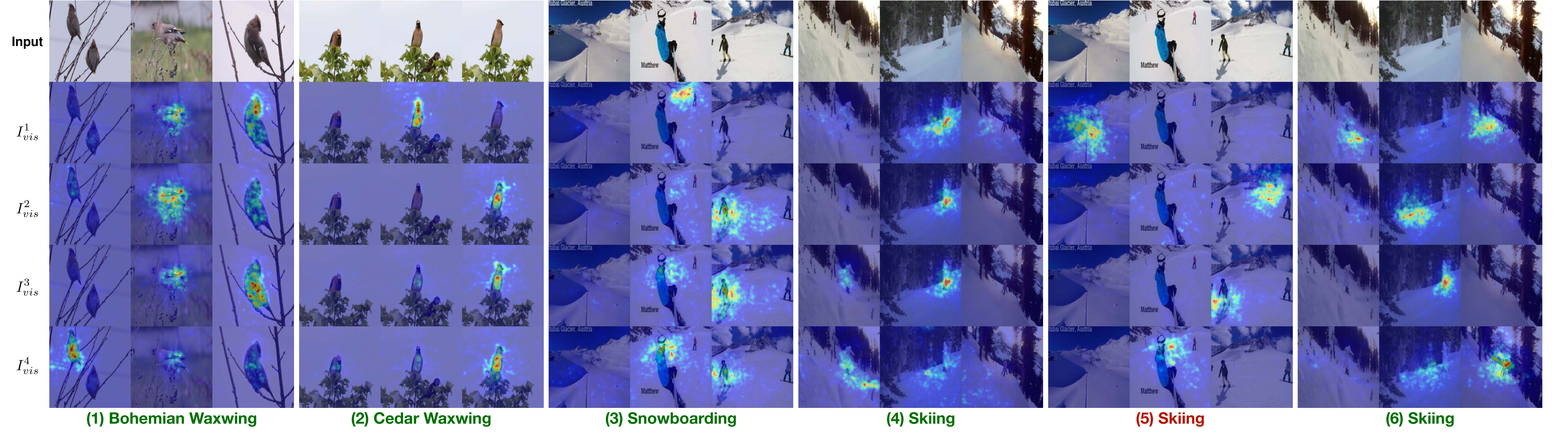}
\caption{
\label{fig:quantitative}
Qualitative results. Red color on heat maps indicate higher attention. (1,2) come from YouTube Birds, the rest come from ActivityNet. Green words are correct answers, red words are wrong answers. The answer of (5) should be SnowBoarding. (1)(2): Results of our model. The 2 birds are very similar, except for their bellies and tails. Our model firstly focus on texture of wings and faces ($I_{vis}^{1}$) to recognize general species, and then colors of bellies ($I_{vis}^{4}$) to distinguish the 2 species. (3,4): Results of our model. The first glimpse/middle two/last glimpse tend to focus on backgrounds/human pose/both background and pose. (5,6): Results of parallel attentions. In (5), all 4 glimpses happen to focus on background and the prediction is wrong since the glimpses are independent.}
\end{figure*}

Fig.~\ref{fig:quantitative} shows qualitative visualizations on YouTube Birds and ActivityNet v1.3 to demonstrate how the attention modules work. The heat maps are drawn with Eq.~\ref{eq:l1}. We select two similar classes for each dataset. Our model attends to the correct region in all cases, while parallel attention fails in one case. The visualizations also demonstrate the complementarity of the glimpses given by our model. In (3,4), its first glimpse tends to be more general, focusing on the surroundings, which is only a weak indicator of actions since both actions are on snow fields. Thanks to the specifically designed redundancy reduction structure, activations of channels representing background features have been weakened after the first iteration. Later glimpses focus more on the human pose, more helpful to identifying activities. However, it is the combination of background and human pose that gives more accurate predictions, so both are attended in the end. Comparing Fig.~\ref{fig:quantitative} (3,4) with (5,6), the advantage of our model is evident. It may happen by chance for the parallel glimpses model that all glimpses focus on the background and being redundant, leading to a wrong prediction. However, in our model, the glimpses can cooperate and get rid of this problem.

\section{Conclusion}
We have demonstrated the Redundancy Reduction Attention (RRA) structure, which aims to extract features of multiple discriminative patterns for fine-grained video categorization. It consists of a spatio-temporal soft attention which summarizes the video, and a suppress-thresholding structure which decreases the redundant activations. Experiments on four video classification datasets demonstrate the effectiveness of the proposed structure. We also release two video datasets for fine-grained categorization, which will be helpful to the community in the future.

%
%
%
%
\clearpage
\bibliographystyle{splncs04}
\bibliography{egbib}
\end{document}